\documentclass[twoside,12pt]{article}
\pdfoutput=1

\usepackage{geometry}
\usepackage{natbib}
\usepackage{graphicx}
\usepackage{rotating}
\usepackage{amsmath}
\usepackage{amssymb}
\usepackage{theorem}
\newcommand{\BlackBox}{\rule{1.5ex}{1.5ex}}  % end of proof

\newcommand{\cbr}[1]{\left\{#1\right\}}

\newcommand{\intset}[1]{\cbr{1..n}}

\usepackage[colorlinks,pdfpagelabels,plainpages=false]{hyperref}
\usepackage{algorithm}
\usepackage{algorithmic}
\usepackage{rotating}

\usepackage{xcolor}
\definecolor{dark-red}{rgb}{0.4,0.15,0.15}
\definecolor{dark-blue}{rgb}{0.15,0.15,0.4}
\definecolor{medium-blue}{rgb}{0,0,0.5}
\hypersetup{
   colorlinks, linkcolor={dark-blue},
   citecolor={dark-blue}, urlcolor={medium-blue}
}

\usepackage{booktabs}
\usepackage{bm}
\usepackage{amsmath}
\usepackage{amssymb}
\usepackage{amsfonts}
\usepackage{mathtools}
\usepackage{comment}
\usepackage{subfigure}

\newcommand{\mbf}[1]{{\boldsymbol{\mathbf{#1}}}}
\renewcommand{\bm}{\mbf}

\usepackage{color}

\usepackage{array}
\newcolumntype{L}[1]{>{\raggedright\let\newline\\\arraybackslash\hspace{0pt}}m{#1}}
\newcolumntype{C}[1]{>{\centering\let\newline\\\arraybackslash\hspace{0pt}}m{#1}}
\newcolumntype{R}[1]{>{\raggedleft\let\newline\\\arraybackslash\hspace{0pt}}m{#1}}

\usepackage{parskip}

\title{Kernel Interpolation for Scalable Structured Gaussian Processes (KISS-GP)}

\author{
  Andrew Gordon Wilson \\
  Carnegie Mellon University \\
  andrewgw@cs.cmu.edu
  \and
  Hannes Nickisch \\
  Philips Research Hamburg \\
  hannes@nickisch.org
}

\begin{document}
\date{}  

\maketitle

\begin{abstract} 
\begin{sloppypar}
We introduce a new \emph{structured kernel interpolation} (SKI) framework, which generalises and 
unifies inducing point methods for scalable Gaussian processes (GPs).  SKI methods produce kernel approximations 
for fast computations through kernel interpolation.  The SKI framework clarifies how the quality of an inducing point 
approach depends on the number of inducing (aka interpolation) points, interpolation strategy, and GP covariance kernel.
SKI also provides a mechanism to create new scalable kernel methods, through choosing different kernel interpolation
strategies.  Using SKI, with local cubic kernel interpolation, we introduce KISS-GP, which is 
1) more scalable 
than inducing point alternatives,  2) naturally enables
Kronecker and Toeplitz algebra for substantial additional gains in scalability, without requiring
any grid data, and 3) can be used for fast and expressive kernel learning.  KISS-GP costs 
$\mathcal{O}(n)$ time and storage for GP inference.
We evaluate KISS-GP for kernel matrix approximation, 
kernel learning, and natural sound modelling.
\end{sloppypar}

\end{abstract} 

\section{Introduction}
\label{sec: intro}

Gaussian processes (GPs) are exactly the types of models we want to apply to big data:
flexible function approximators, capable of using the information in large datasets to learn intricate
structure through interpretable and expressive covariance kernels.  However, $\mathcal{O}(n^3)$ 
and $\mathcal{O}(n^2)$  computation and storage requirements limit GPs to all but the smallest 
datasets, containing at most a few thousand training points $n$.  Their impressive empirical successes thus 
far 
are only a glimpse of what might be possible, if only we could overcome these computational 
limitations \citep{rasmussenphd96}.

\begin{sloppypar}
Inducing point methods \citep{snelson2006sparse, hensman2013uai, candela2005, seeger2005thesis, 
smola2001sparse, silverman1985} have been introduced to scale up Gaussian processes to larger 
datasizes.
These methods cost 
$\mathcal{O}(m^2 n + m^3)$ computations and $\mathcal{O}(mn + m^2)$ storage, for $m$ inducing points,
and $n$ training data points.  Inducing methods are popular for their general purpose ``out of the box'' 
applicability, without requiring any special structure in the data.  However, these methods are limited by
requiring a small $m \ll n$ number of inducing inputs, which can cause a deterioration in predictive
performance, and the inability to perform expressive kernel learning \citep{wilsonkernel2014}.
\end{sloppypar}

Structure exploiting approaches for scalability, such as Kronecker \citep{saatchi11} or 
Toeplitz \citep{cunningham2008fast} methods, have orthogonal advantages to inducing point methods.  
These methods exploit the existing structure in the covariance kernel for highly 
accurate and scalable inference, and can be used for flexible kernel learning on large datasets \citep{wilsonkernel2014}.
However, Kronecker methods require that inputs (predictors) are on a multidimensional lattice (a Cartesian product grid), 
which makes them inapplicable to most datasets.  Although \citet{wilsonkernel2014} has extended Kronecker methods for partial grid 
structure, these extensions do not apply to arbitrarily located inputs.  
Likewise, the Kronecker based approach in \citet{luo2013fast} 
involves costly rank-1 updates and is not generally
applicable for arbitrarily located inputs.  Toeplitz methods are similarly restrictive, requiring that the data
are on a regularly spaced 1D grid.

It is tempting to assume we could place inducing points on a grid, and then take advantage of 
Kronecker or Toeplitz structure for further gains in scalability.  However, this naive approach 
only helps reduce the $m^3$ complexity term in inducing point methods, and not the more critical 
$m^2 n$ term, which arises from a matrix of cross covariances between training and inducing inputs.

In this paper, we introduce a new unifying framework for inducing point methods, called 
\emph{structured kernel interpolation} (SKI).  This framework allows us to improve the 
scalability and accuracy of fast kernel methods, 
and to naturally combine the advantages
of inducing point and structure exploiting approaches.  In particular, 
\begin{itemize}
\item We show how current inducing point methods can be interpreted as performing
a global GP interpolation on a true underlying kernel to create an approximate kernel for 
scalable computations, as part of a more general family of \emph{structured kernel 
interpolation} methods.
\item The SKI framework helps us understand how the accuracy and efficiency of an inducing 
point method is affected by the number of inducing points $m$, the choice of kernel, and the 
choice of interpolation method.  Moreover, by choosing different interpolation strategies for SKI,
we can create new inducing point methods.
\item We introduce a new inducing point method, KISS-GP, which uses local cubic and inverse distance
weighting interpolation strategies to create a sparse approximation to the cross covariance matrix 
between the inducing points and original training points.  This method can naturally be 
combined with Kronecker and Toeplitz algebra to allow for $m \gg n$ inducing points,
and further gains in scalability.  When exploiting Toeplitz structure KISS-GP requires 
$\mathcal{O}(n+m \log m)$ computations and $\mathcal{O}(n+m)$ 
storage.  When exploiting Kronecker structure, KISS-GP requires 
$\mathcal{O}(n+Pm^{1+1/P})$ computations and $\mathcal{O}(n+Pm^{2/P})$ storage,
for $P > 1$ dimensional inputs.
\item KISS-GP can be viewed as lifting the grid restrictions in Toeplitz and Kronecker methods, 
so that one can use arbitrarily located inputs.
\item We show that the ability for KISS-GP to efficiently use a large number of inducing points 
enables expressive kernel learning, and orders of magnitude greater accuracy and efficiency
over popular alternatives such as FITC \citep{snelson2006sparse}.
\item We have implemented code as an extension to the GPML toolbox \citep{rasmussen10gpml}.
\item Overall, the simplicity and generality of the SKI framework makes it easy to 
design scalable Gaussian process methods with high accuracy and low computational costs.
\end{itemize}

We start in section \ref{sec: background} with background on Gaussian processes
(section \ref{sec: gps}), inducing point methods (section \ref{sec: inducing}), and structure exploiting methods (section \ref{sec: structure}). 
We then introduce the structured kernel interpolation (SKI) framework, and the KISS-GP method, in section \ref{sec: SKI}.  
In section \ref{sec: experiments} we conduct experiments
on kernel matrix reconstruction, kernel learning, and natural sound modelling.  We conclude in section \ref{sec: discussion}.  

\section{Background}
\label{sec: background}

\subsection{Gaussian Processes}
\label{sec: gps}

We provide a brief review of Gaussian processes \citep{rasmussen06}, and 
the associated computational requirements for inference and learning.
Throughout we assume we have a dataset $\mathcal{D}$ of $n$ 
input (predictor) vectors $X = \{\bm{x}_1,\dots,\bm{x}_n\}$, each of dimension $D$,
corresponding to a $n \times 1$ vector of targets 
$\bm{y} = (y(\bm{x}_1),\dots,y(\bm{x}_n))^{\top}$.

A Gaussian process (GP) is a collection of random variables, any finite number of which have a joint Gaussian distribution.
Using a GP, we can define a distribution over functions $f(\bm{x}) \sim \mathcal{GP}(\mu,k)$, meaning that any 
collection of function values $\bm{f}$ has a joint Gaussian distribution:
\begin{align}
 \bm{f} = f(X) = [f(\bm{x}_1),\dots,f(\bm{x}_n)]^{\top} \sim \mathcal{N}(\bm{\mu},K) \,.  \label{eqn: gpdef}
\end{align}
The $n \times 1$ mean vector $\bm{\mu}_i = \mu(\bm{x}_i)$, and $n \times n$ covariance matrix $K_{ij} = k(\bm{x}_i,\bm{x}_j)$,
are defined by the user specified mean function $\mu(\bm{x}) = \mathbb{E}[f(\bm{x})]$ and covariance kernel $k(\bm{x},\bm{x}') = \text{cov}(f(\bm{x}),f(\bm{x}'))$ 
of the Gaussian process.  The smoothness and generalisation properties of the GP are encoded by the 
covariance kernel and its hyperparameters $\bm{\theta}$.  For example, the popular RBF
covariance function, with length-scale
hyperparameter $\ell$, has the form
\begin{equation}
 k_{\text{RBF}}(\bm{x},\bm{x}') = \text{exp}(-0.5 ||\bm{x}-\bm{x}'||^2/\ell^2) \,. \label{eqn: rbfcov}
\end{equation}

If the targets $y(\bm{x})$ are modelled by a GP with additive 
Gaussian noise, e.g., $y(\bm{x})|f(\bm{x}) \sim \mathcal{N}(y(\bm{x}); f(\bm{x}),\sigma^2)$,
the predictive distribution at $n_*$ test points $X_*$ is given by
\begin{align}
 \bm{f}_*|X_*,&X,\bm{y},\bm{\theta},\sigma^2 \sim \mathcal{N}(\bar{\bm{f}}_*,\text{cov}(\bm{f}_*)) \,, \label{eqn: fullpred}  \\  
 \bar{\bm{f}}_* &= \bm{\mu}_{X_*}  + K_{X_*,X}[K_{X,X}+\sigma^2 I]^{-1}\bm{y}\,,   \notag \\ 
 \text{cov}(\bm{f}_*) &= K_{X_*,X_*} - K_{X_*,X}[K_{X,X}+\sigma^2 I]^{-1}K_{X,X_*} \,.  \notag
\end{align}
$K_{X_*,X}$, for example, denotes the $n_* \times n$ matrix of covariances between the GP evaluated 
at $X_*$ and $X$. $\bm{\mu}_{X_*}$ is the $n_* \times 1$ mean vector,
and $K_{X,X}$ is the $n \times n$ covariance matrix evaluated at training inputs $X$.
All covariance matrices implicitly depend on the kernel hyperparameters $\bm{\theta}$.

We can analytically marginalise the Gaussian process $f(\bm{x})$ to obtain the 
marginal likelihood of the data, conditioned only on the covariance hyperparameters $\bm{\theta}$:
\begin{equation}
 \log p(\bm{y}|\bm{\theta}) \propto -[\overbrace{\bm{y}^{\top}(K_{\bm{\theta}}+\sigma^2 I)^{-1}\bm{y}}^{\text{model fit}} + \overbrace{\log|K_{\bm{\theta}} + \sigma^2 I|}^{\text{complexity penalty}}]\,.  \label{eqn: mlikeli}
\end{equation}
Eq.~\eqref{eqn: mlikeli} nicely separates into automatically calibrated 
model fit and complexity terms \citep{rasmussen01}, and can be optimized to learn the kernel
hyperparameters $\bm{\theta}$, or used to integrate out $\bm{\theta}$ via MCMC \citep{rasmussenphd96}.  

The computational bottleneck in using Gaussian processes is solving a linear system
$(K+\sigma^2 I)^{-1}\bm{y}$ (for inference), and $\log |K + \sigma^2 I|$ (for hyperparameter learning).  
For this purpose, standard procedure is to compute the Cholesky decomposition of $K$, requiring 
$\mathcal{O}(n^3)$ operations and $\mathcal{O}(n^2)$ storage.  Afterwards, the predictive mean
and variance respectively cost $\mathcal{O}(n)$ and $\mathcal{O}(n^2)$ for a single test point $\mathbf{x}_*$.

\subsection{Inducing Point Based Sparse Approximations}
\label{sec: inducing}

Many popular approaches to scaling up GP inference belong to a family of 
inducing point methods \citep{candela2005}.  These methods can be viewed 
as replacing the exact kernel $k(\mathbf{x}, \mathbf{z})$ by an approximation
$\tilde{k} (\mathbf{x},\mathbf{z})$ for fast computations.

For example, the prominent subset of regressors (SoR) 
\citep{silverman1985} and fully independent training conditional (FITC) 
\citep{snelson2006sparse} methods use the approximate kernels
\begin{align}
  \tilde{k}_{\text{SoR}} (\mathbf{x}, \mathbf{z}) &= K_{\mathbf{x},U} K_{U,U}^{- 1} K_{U,\mathbf{z}} \label{eqn: ksor} \,,  \\
  \tilde{k}_{\text{FITC}} (\mathbf{x}, \mathbf{z}) &= \tilde{k}_{\text{SoR}} (\mathbf{x}, \mathbf{z}) 
  									    + \delta_{\bm{x}\bm{z}} \left(k(\mathbf{x}, \mathbf{z}) - \tilde{k}_{\text{SoR}} (\mathbf{x}, \mathbf{z}) \right) \,, \label{eqn: kfitc} 
\end{align}
for a set of $m$ inducing points $U = [\mathbf{u}_i]_{i=1 \dots m}$.  $K_{\mathbf{x},U}$, $K_{U,U}^{- 1}$, and $K_{U,\mathbf{z}}$ are generated
from the exact kernel $k(\mathbf{x}, \mathbf{z})$.
While SoR yields an $n \times n$ covariance matrix $K_{\text{SoR}}$ of 
rank at most $m$, corresponding to a degenerate (finite basis) Gaussian process, FITC leads to a full rank covariance 
matrix $K_{\text{FITC}}$ due to its diagonal correction.  As a result, FITC is a more faithful approximation and is 
preferred in practice.  Note that the exact user-specified kernel, $k(\mathbf{x}, \mathbf{z})$, will be parametrized
by $\bm{\theta}$, and therefore kernel learning in an inducing point method takes place by, e.g.,\ optimizing the 
SoR or FITC marginal likelihoods with respect to $\bm{\theta}$.

These approximate kernels give rise to $\mathcal{O}(m^2 n + m^3)$ computations and $\mathcal{O}(mn + m^2)$ storage for GP inference and 
learning \citep{candela2005}, after which the GP predictive mean and variance cost $\mathcal{O}(m)$ and 
$\mathcal{O}(m^2)$ per test case. To see practical efficiency gains over standard inference procedures, one 
is constrained to choose $m \ll n$, which often leads to a severe deterioration in predictive performance, and 
an inability to perform expressive kernel learning \citep{wilsonkernel2014}.

\subsection{Fast Structure Exploiting Inference}
\label{sec: structure}

Kronecker and Toeplitz methods exploit the \emph{existing} structure of the 
GP covariance matrix $K$ to scale up inference and learning without approximations.

\subsubsection{Kronecker Methods} 
\label{sec: kronecker}

We briefly review Kronecker methods.  A full introduction is provided in chapter 5 of \citet{saatchi11}.

If we have multidimensional inputs on a Cartesian grid, $\bm{x} \in \mathcal{X}_1 \times \dots \times \mathcal{X}_P$,
and a product kernel across grid dimensions, $k(\bm{x}_i,\bm{x}_j) = \prod_{p=1}^{P} k(\bm{x}_i^{(p)},\bm{x}_j^{(p)})$, 
then the $m \times m$ covariance matrix $K$ can be expressed as a Kronecker product $K = K_1 \otimes \dots \otimes K_P$
(the number of grid points $m = \prod_{i=1}^{P} n_p$ is a product of the number of points $n_p$ per grid dimension).
It follows that we can efficiently find the eigendecomposition of $K = Q V Q^{\top}$ by separately computing the 
eigendecomposition of each of $K_1,\dots,K_P$.  One can similarly exploit Kronecker structure for fast matrix 
vector products \citep{wilsonkernel2014}.  

Fast eigendecompositions and matrix vector products of Kronecker matrices allow us to efficiently
evaluate $(K+\sigma^2 I)^{-1} \bm{y}$ and $\log |K+\sigma^2 I |$ for scalable and exact inference
and learning with GPs.  Specifically, given an eigendecomposition 
of $K$ as $QVQ^{\top}$, we can write $(K+\sigma^2 I)^{-1}\bm{y} = (QVQ^{\top} + \sigma^2 I)^{-1}\bm{y} = Q(V+\sigma^2 I)^{-1} Q^{\top} \bm{y}$,
and $\log |K + \sigma^2 I | = \sum_i \log(V_{ii} + \sigma^2)$.  $V$ is a diagonal matrix of eigenvalues,
so inversion is trivial.  $Q$, an orthogonal matrix of eigenvectors, also decomposes as a Kronecker
product, which enables fast matrix vector products.  Overall, inference and learning cost 
$\mathcal{O}(Pm^{1+1/P})$ operations (for $P>1$) and $\mathcal{O}(Pm^{\frac{2}{P}})$ storage 
\citep{saatchi11, wilsonkernel2014}.  

While product kernels can be easily constructed, and popular kernels such as the RBF kernel of 
Eq.~\eqref{eqn: rbfcov} already have product structure, requiring a multidimensional input grid 
can be a severe constraint.  

\citet{wilsonkernel2014} extend Kronecker methods to datasets with only partial grid 
structure -- e.g.,\ images with random missing pixels, or spatiotemporal grids with missing data due to 
water.   They complete a partial grid with virtual observations, and use a diagonal noise covariance matrix $A$ 
which ignores the effects of these virtual observations: $K^{(n)} + \sigma^2 I \to K^{(m)} + A$, 
where $K^{(n)}$ is an $n \times n$ covariance matrix formed from the original dataset with $n$ datapoints,
and $K^{(m)}$ is the covariance matrix after augmentation from virtual inputs.
Although we cannot efficiently eigendecompose 
$K^{(m)}+A$, we can take matrix vector products $(K^{(m)} + A)\bm{y}^{(m)}$ efficiently, since $K^{(m)}$ is Kronecker and $A$ is diagonal.  
We can thus compute $(K^{(m)}+A)^{-1}\bm{y}^{(m)} = (K^{(n)}+\sigma^2 I)^{-1}\bm{y}^{(n)}$ to within machine precision, and perform efficient inference, using iterative methods such as linear conjugate 
gradients, which only involve matrix vector products.

To evaluate the marginal likelihood in Eq.~\eqref{eqn: mlikeli}, for kernel learning, we must also 
compute $\log |K^{(n)} + \sigma^2 I|$, where $K^{(n)}$ is an $n \times n$ covariance matrix 
formed from the original dataset with $n$ datapoints.  \citet{wilsonkernel2014} 
propose to approximate the eigenvalues $\lambda_i^{(n)}$ of $K^{(n)}$ using the largest $n$ eigenvalues 
$\lambda_i$ of $K^{(m)}$, the Kronecker covariance matrix formed from the completed grid,
which can be eigendecomposed efficiently.  In particular,
\begin{align}
\log |K^{(n)} + \sigma^2 I | &= \sum_{i=1}^{n} \log ({\lambda}_i^{(n)} + \sigma^2)  
                                         \approx \sum_{i=1}^{n} \log(\frac{n}{m}{\lambda}_i + \sigma^2) \,. \notag
\end{align}
Theorem 3.4 of \citet{baker1977numerical} proves this eigenvalue approximation is asymptotically consistent (e.g., converges in the
limit of large $n$), so long as the observed inputs are bounded by the complete grid. \citet{williams2003stability} also show 
that one can bound the true eigenvalues by their approximation using PCA. 
Notably, only the log determinant (complexity penalty) term in the marginal likelihood
undergoes a small approximation.  \citet{wilsonkernel2014} show that, in practice, this approximation can be 
highly effective for fast and expressive kernel learning.

However, the extensions in \citet{wilsonkernel2014} are only efficient if the input space has partial 
grid structure, and do not apply in general settings.

\subsubsection{Toeplitz Methods}
\label{sec: toeplitz}

Toeplitz and Kronecker methods are complementary.  $K$ is a Toeplitz covariance 
matrix if it is generated from a stationary covariance kernel, $k(\bm{x},\bm{x}') = k(\bm{x}-\bm{x}')$, with inputs $\bm{x}$ on a regularly spaced one dimensional grid.  Toeplitz matrices are constant along their diagonals: $K_{i,j} = K_{i+1,j+1} = k(\bm{x}_i - \bm{x}_j)$.   

One can embed Toeplitz matrices into circulant matrices, to perform fast matrix vector
products using fast Fourier transforms, e.g., \citet{wilson2014thesis}.  One can then 
use linear conjugate gradients to solve linear systems $(K+\sigma^2 I)^{-1}\bm{y}$ in 
$\mathcal{O}(m \log m)$ operations and $\mathcal{O}(m)$ storage, for $m$ 
grid datapoints.  \citet{turner10} and \citet{cunningham2008fast} contain examples of 
Toeplitz methods applied to GPs.

\section{Structured Kernel Interpolation}
\label{sec: SKI}

We wish to ease the large $\mathcal{O}(n^3)$ computations
and $\mathcal{O}(n^2)$ storage associated with Gaussian 
processes, while retaining model flexibility and general 
applicability.

Inducing point approaches (section \ref{sec: inducing}) to scalability are popular
because they can be applied ``out of the box'', 
without requiring special structure in the data.  However, with a small
number of inducing points, these methods suffer from a major deterioration 
in predictive accuracy, and the inability to perform expressive kernel learning 
\citep{wilsonkernel2014}, which will be most valuable on large datasets.
On the other hand, structure exploiting approaches (section \ref{sec: structure}) are
compelling because they provide incredible gains in scalability, with essentially
no losses in predictive accuracy.  But the requirement of an input grid makes 
these methods inapplicable to most problems.

Looking at equations \eqref{eqn: ksor} and \eqref{eqn: kfitc}, it is tempting to
try placing the locations of the inducing points $U$ on a grid, in the SoR or 
FITC methods, and then exploit either Kronecker or Toeplitz algebra to efficiently 
solve linear systems involving $K_{U,U}^{-1}$.  While this naive approach would 
reduce the $\mathcal{O}(m^3)$ complexity associated with $K_{U,U}^{-1}$, that
is not the dominant term for computations with inducing point methods -- rather, 
it is the $\mathcal{O}(m^2 n)$ computations associated with the product 
$K_{X,U} K_{U,U}^{-1}$.  

We observe, however, that we can approximate the $n \times m$ matrix $K_{X,U}$ 
of cross covariances for the kernel evaluated at the training and inducing inputs 
$X$ and $U$, by interpolating on the $m \times m$ covariance matrix $K_{U,U}$.
For example, if we wish to estimate $k(\bm{x}_i,\bm{u}_j)$, for input point 
$\bm{x}_i$ and inducing point $\bm{u}_j$, we can start by finding the two inducing
points $\bm{u}_a$ and $\bm{u}_b$ which most closely bound $\bm{x}_i$:
$\bm{u}_a \leq \bm{x}_i \leq \bm{u}_b$ (initially assuming $D=1$ and a Toeplitz $K_{U,U}$ 
from a regular grid $U$, for simplicity).  
We can then form 
$\tilde{k}(\bm{x}_i,\bm{u}_j) = w_i k(\bm{u}_a,\bm{u}_j) + (1-w_i) k(\bm{u}_b,\bm{u}_j)$,
with linear interpolation weights $w_i$ and $(1-w_i)$, which represent the relative distances from 
$\bm{x}_i$ to points $\bm{u}_a$ and $\bm{u}_b$.  More generally, we form 
\begin{align}
{K}_{X,U} \approx W K_{U,U} \,,  \label{eqn: crossinterp}
\end{align}
where $W$ is an $n \times m$ matrix of interpolation weights.  We observe that $W$ can be extremely 
sparse.  For local linear interpolation, $W$ contains only $c=2$ non-zero entries per row -- the interpolation
weights -- which sum to $1$.  
For greater accuracy, we 
can use local cubic interpolation \citep{keys1981}
on equispaced grids, 
in which case $W$ has $c=4$ non-zero entries per row.  For general rectilinear 
grids $U$ (without regular spacing), we can use inverse distance weighting \citep{shepard1968} with $c=2$ non-zero weights
per row of $W$.  

Substituting our expression for $\tilde{K}_{X,U}$ in Eq.~\eqref{eqn: crossinterp} into the SoR approximation 
for $K_{X,X}$, we find:
\begin{align}
K_{X,X}  &\overset{\text{SoR}} \approx K_{X,U} K_{U,U}^{-1} K_{U,X}   \overset{\text{Eq.~\eqref{eqn: crossinterp}}}\approx  W K_{U,U} K_{U,U}^{-1} K_{U,U} W^{\top} \notag \\ 
&= W K_{U,U} W^{\top} = K_{\text{SKI}} \,.   \label{eqn: kski}
\end{align}
We name this general approach to approximating GP kernel functions   
\emph{structured kernel interpolation} (SKI).  Although we have made
use of the SoR approximation as an example, SKI can be applied to essentially
any inducing point method, such as FITC.\footnote{Combining with the SoR approximate $k(x,z)$,
one can naively use $k_{\text{SKI}}(\bm{x},\bm{z}) = \bm{w}_x^{\top} K_{U,U} \bm{w}_z$,
where $\bm{w}_x,\bm{w}_z \in \mathbf{R}^m$ are interpolation vectors for points
$\bm{x}$ and $\bm{z}$; however, when $\bm{w}_x \neq \bm{w}_z$, 
it makes most sense to perform local interpolation on $K_{U,\bm{z}}$ directly.}$^{,}$\footnote{Later 
we discuss the logistics of combining with FITC.}

We can compute fast matrix vector products $K_{\text{SKI}} \bm{y}$.
If we do not exploit Toeplitz or Kronecker structure in $K_{U,U}$, a matrix
vector product costs $\mathcal{O}(n+m^2)$ computations and $\mathcal{O}(n+m^2)$
storage (matrix vector products with $K_{U,U}$ cost $\mathcal{O}(m^2)$ computations, and with
sparse $W$ cost $\mathcal{O}(n)$, with the same storage requirements).
If we exploit Kronecker structure, we only require 
$\mathcal{O}(Pm^{1+1/P})$ computations and $\mathcal{O}(n+Pm^{\frac{2}{P}})$ storage
(matrix vector products with $K_{U,U}$ now cost $\mathcal{O}(Pm^{1+1/P})$ computations and
$\mathcal{O}(Pm^{\frac{2}{P}})$ storage).
If we exploit Toeplitz structure, we only require 
$\mathcal{O}(n + m \log m)$ computations and $\mathcal{O}(n+m)$ storage
(a matrix vector product with $K_{U,U}$ now costs $\mathcal{O}(m \log m)$ computations
and $\mathcal{O}(m)$ storage).  

Inference proceeds by solving $K_{\text{SKI}}^{-1} \bm{y}$ through linear conjugate gradients, which
only requires matrix vector products and a small number $j \ll n$ of iterations for convergence
to within machine precision.  To compute $\log |K_{\text{SKI}}|$, for the marginal likelihood evaluations 
used in kernel learning, one can follow the approximation of \citet{wilsonkernel2014}, described in 
section \ref{sec: kronecker}, where $K_{U,U}$ takes the role of $K^{(m)}$, and virtual 
observations are not required.  Alternatively, we can use the ability to take fast matrix vector
products with $K_{\text{SKI}}$ in standard eigenvalue solvers to efficiently compute the log
determinant exactly.  We can also form an accurate approximation by selectively computing the
largest and smallest eigenvalues.  This alternative approach is not possible in 
\citet{wilsonkernel2014} since in that case one cannot take fast matrix vector products with $K^{(n)}$.
In either fast approach, the computional complexity for learning
is no greater than for inference.

In short, even if we choose \emph{not} to exploit potential Kronecker or 
Toeplitz structure in $K_{U,U}$, inference and learning in SKI are 
accelerated over standard inducing point approaches.  However, unlike
with the data inputs, $X$, which are fixed, we are free to choose the locations
of the latent inducing points $U$, and therefore we can easily create 
(e.g.,\ Toeplitz or Kronecker) structure in $K_{U,U}$ which might not exist 
in $K_{X,X}$.  In the SKI formalism, we can 
uniquely exploit this structure for substantial additional gains in efficiency, and the ability
to use an unprecedented number of inducing points, while lifting any grid requirements 
for the training inputs $X$.

Although here we have made use of the SoR approximation in Eq.~\eqref{eqn: kski}, 
we could trivially apply the FITC diagonal correction (section \ref{sec: inducing}), or 
combine with other approaches.  However, within the SKI framework, the diagonal
correction of FITC does not have as much value: $K_{\text{SKI}}$ can easily be full
rank and still have major computational benefits, using $m > n$.  In conventional 
inducing approximations, one would never set $m > n$, since this would be less 
efficient than exact Gaussian process inference.

Finally, we can understand all of these inducing approaches as part of a general 
\emph{structured kernel interpolation} (SKI) framework.  
The predictive mean $\bar{f}_{*}$ in Eq.~(\ref{eqn: fullpred}) of a noise-free,
zero mean GP ($\sigma=0,\: \mu(\mathbf{x})\equiv0$) is linear in two
ways: on the one hand, as a $\mathbf{w}_{X}(\mathbf{x}_{*})=K_{X,X}^{-1}K_{X,\mathbf{x}_{*}}$
weighted sum of the observations $\mathbf{y}$, and on the other hand
as an $\bm{\alpha}=K_{X,X}^{-1}\mathbf{y}$ weighted sum of training-test
cross-covariances $K_{X,\mathbf{x}_{*}}$:
\begin{align}
  \bar{f}_{*}=\mathbf{y}^{\top}\mathbf{w}_{X}(\mathbf{x}_{*})=\bm{\alpha}^{\top}K_{X,\mathbf{x}_{*}} \,.
\end{align}
If we are to perform a noise free zero-mean GP regression on the kernel itself, such that we have data
$\mathcal{D} = (\mathbf{u}_{i},k(\mathbf{u}_{i},\mathbf{x}))_{i=1}^{m}$,
then we recover the SoR kernel $\tilde{k}_{\text{SoR}}(\bm{x},\bm{z})$ of equation \eqref{eqn: ksor} as the 
predictive mean of the GP at test point $\bm{x}_* = \bm{z}$.   This finding provides a new unifying perspective 
on inducing point approaches:  all conventional inducing point methods, such as SoR and FITC, can be 
re-derived as performing a zero-mean Gaussian process interpolation on the true kernel.  Indeed, we could write 
\emph{interpolation points} instead of \emph{inducing points}.  The $n \times m$ interpolation weight matrix $W$, 
in all conventional cases, will have all non-zero entries, which incurs great computational expenses.  And, computional 
considerations aside, it is not ideal for inducing point methods to perform a zero-mean GP regression on a covariance 
kernel.  For example, since popular kernels are often strictly positive -- neither zero-mean, nor accurately characterized 
by a Gaussian distribution -- conventional inducing point methods will tend to underestimate covariances.

\begin{figure}[!ht]
\centering
\subfigure[Global Kernel Interpolation]{\label{fig: kone}\includegraphics[scale=0.35]{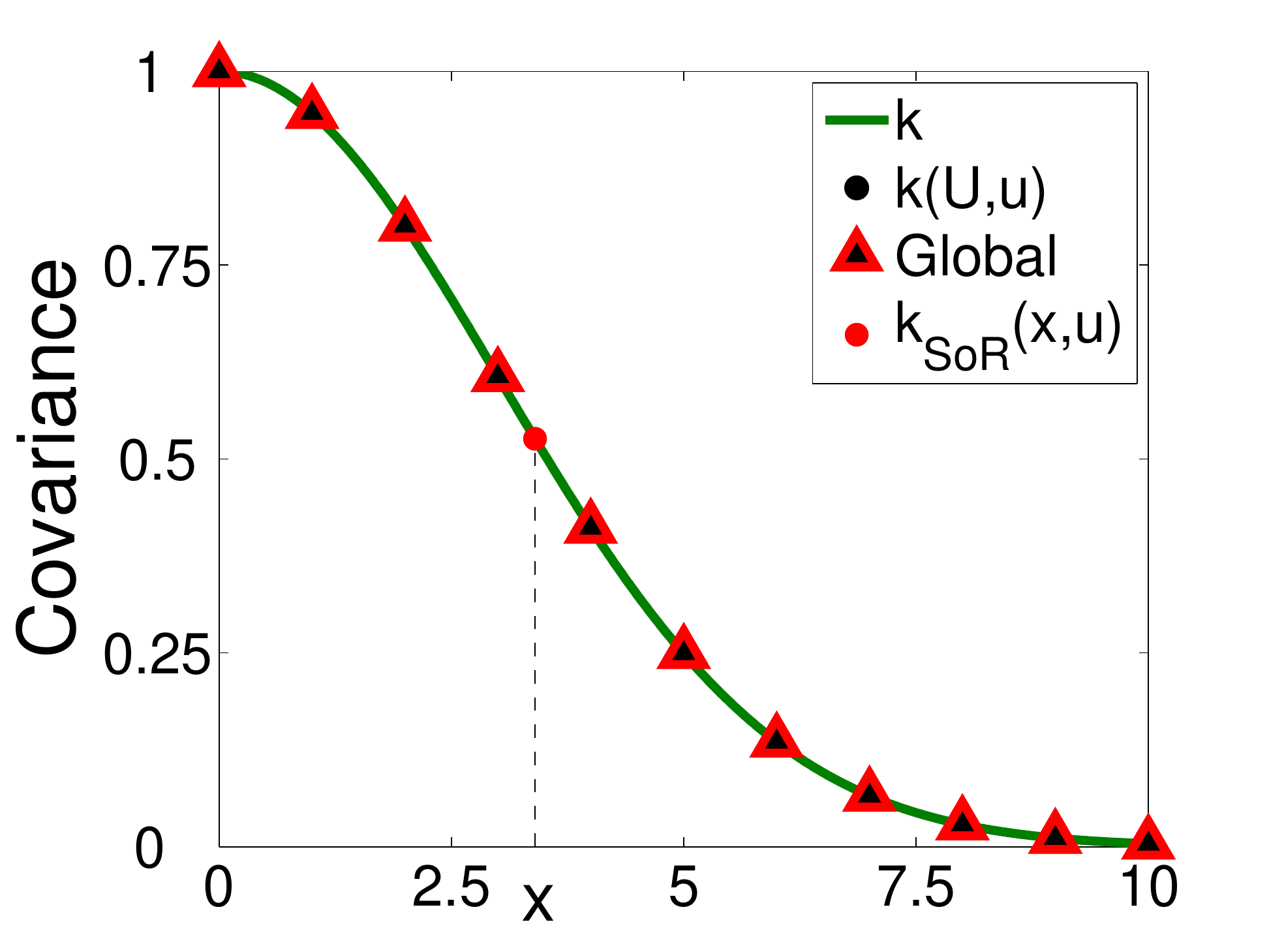}} 
\subfigure[Local Kernel Interpolation]{\label{fig: ktwo}\includegraphics[scale=0.35]{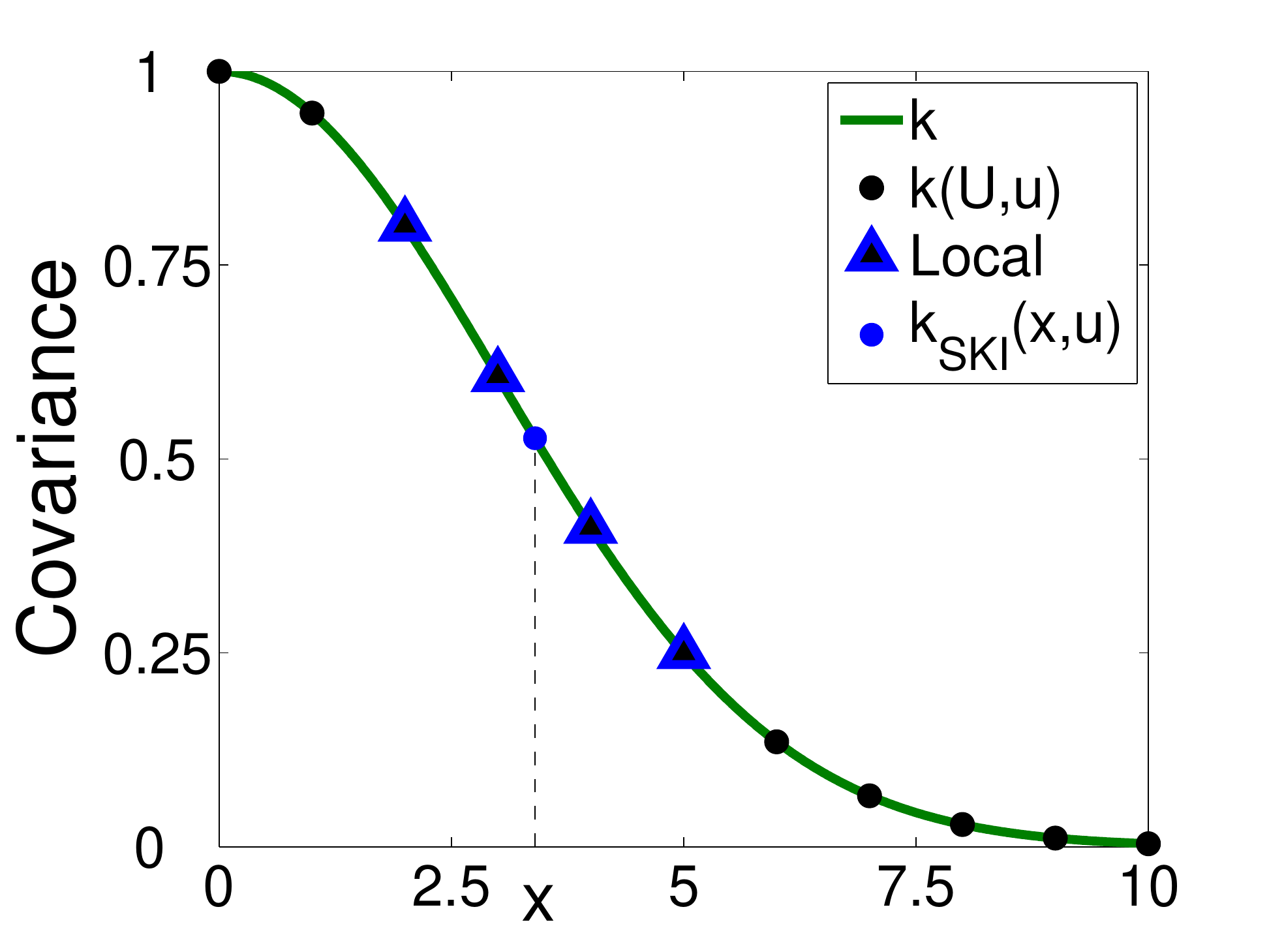}}
\caption{Global vs.\ local kernel interpolation.  Triangle markers denote the inducing points used
for interpolating $k(x,u)$ from $k(U,u)$.  Here $u=0$, $U = \{0,1,\dots,10\}$, and $x=3.4$. a) All conventional inducing 
point methods, such as SoR or FITC, perform global GP regression on $K_{U,u}$ (a vector of covariances 
between all inducing points $U$ and the point $u$), at test point $x_* = x$, to form an approximate 
$\tilde{k}$, e.g., $k_{\text{SoR}}(x,u) = K_{x,U}K_{U,U}^{-1}K_{U,u}$, for any desired $x$ and $u$.  
b) SKI can perform local kernel interpolation on 
$K_{U,u}$ to form the approximation $k_{\text{SKI}}(x,u) = \bm{w}_x^{\top} K_{U,u}$.}
\label{fig: localglobal}
\end{figure}

The SKI interpretation of inducing point methods provides a mechanism to create new inducing
point approaches.  By replacing \emph{global} GP kernel interpolation with \emph{local} inverse distance weighting or 
cubic interpolation, as part of our SKI framework, we make $W$ extremely sparse.  We illustrate the differences 
between local and global kernel interpolation in Figure~\ref{fig: localglobal}.
In addition to the sparsity in $W$, this interpolation 
perspective naturally enables us to exploit (e.g., Toeplitz or Kronecker) structure in the kernel for 
further gains in scalability, without requiring that the inputs $X$ (which index the targets $\bm{y}$) 
are on a grid.

This unifying perspective of inducing methods as kernel interpolation also clarifies when 
these approaches will perform best.  The key assumption, in all of these approaches,
is smoothness in the true underlying kernel $k$.  We can expect interpolation approaches to work well
on popular kernels, such as the RBF kernel, which is a simple exponential function.  More expressive kernels,
such as the spectral mixture kernel \citep{wilsonadams2013}, will require more inducing (interpolation)
points for a good approximation, due to their quasi-periodic nature.  It is our contention that the 
loss in accuracy going from, e.g., global GP kernel interpolation to local cubic kernel interpolation is more than 
recovered by the subsequent ability to greatly increase the number of inducing points.  Moreover, we believe
the structure of most popular kernel functions is conducive to \emph{local} versus \emph{global} interpolation,
resulting in a strong approximation with greatly improved scalability.

When combining SKI with i) GPs, ii) sparse (e.g.\ cubic) interpolation, and iii)
Kronecker or Toeplitz algebra, we name the resulting method KISS-GP, 
though in the experiments of section \ref{sec: experiments}
we will typically write, e.g., ``SKI with cubic interpolation''.  We also use the 
terms \emph{inducing points} and \emph{interpolation points} interchangeably.

\section{Experiments}
\label{sec: experiments}

We evaluate SKI for kernel matrix approximation (section \ref{sec: covrecon}), kernel learning
(section \ref{sec: klearn}), and natural sound modelling (section \ref{sec: natsound}).

We particularly compare with FITC \citep{snelson2006sparse}, because 1) FITC is the most popular inducing point
approach, 2) FITC has been shown to have superior predictive performance and similar efficiency to other inducing 
methods, and is generally recommended \citep{naish2007, quinonero2007},
and 3) FITC is well understood, 
and thus FITC comparisons help elucidate the fundamental properties of SKI,
which is our primary goal.  However, we also provide comparisons with SoR,
and SSGPR \citep{lazaro2010sparse}, 
a recent state of the art scalable GP method based on random projections
with $\mathcal{O}(m^2 n)$ computations and $\mathcal{O}(m^2)$ 
storage for $m$ basis functions and $n$ training points
\citep[see also][]{Rahimi07, le2013fastfood, yang2015carte, LuMayLiuGaretal14}.

Furthermore, we focus on the ability for SKI to allow a relaxation
of Kronecker and Toeplitz methods to arbitrarily located inputs.  
Since Toeplitz methods are restricted to $1$D inputs, and Kronecker methods 
can only be used for low dimensional $(D < 5)$ input spaces \citep{saatchi11},
we consider lower dimensional problems.

All experiments were performed on a 2011 MacBook Pro, with an Intel i5 2.3 GHz processor and 
4 GB of RAM.

\subsection{Covariance Matrix Reconstruction}
\label{sec: covrecon}

Accurate inference and learning depends on the GP covariance matrix $K$, which is used to 
form the predictive distribution and marginal likelihood of a Gaussian process.  We evaluate the 
SKI approximation to $K$, in Eq.~\eqref{eqn: kski}, as a function of number of inducing points $m$, 
inducing point locations, and sparse interpolation strategy.

\begin{figure}[!ht]
\centering
\subfigure[$K_\text{true}$]{\label{fig:ktrue}\includegraphics[scale = 0.25]{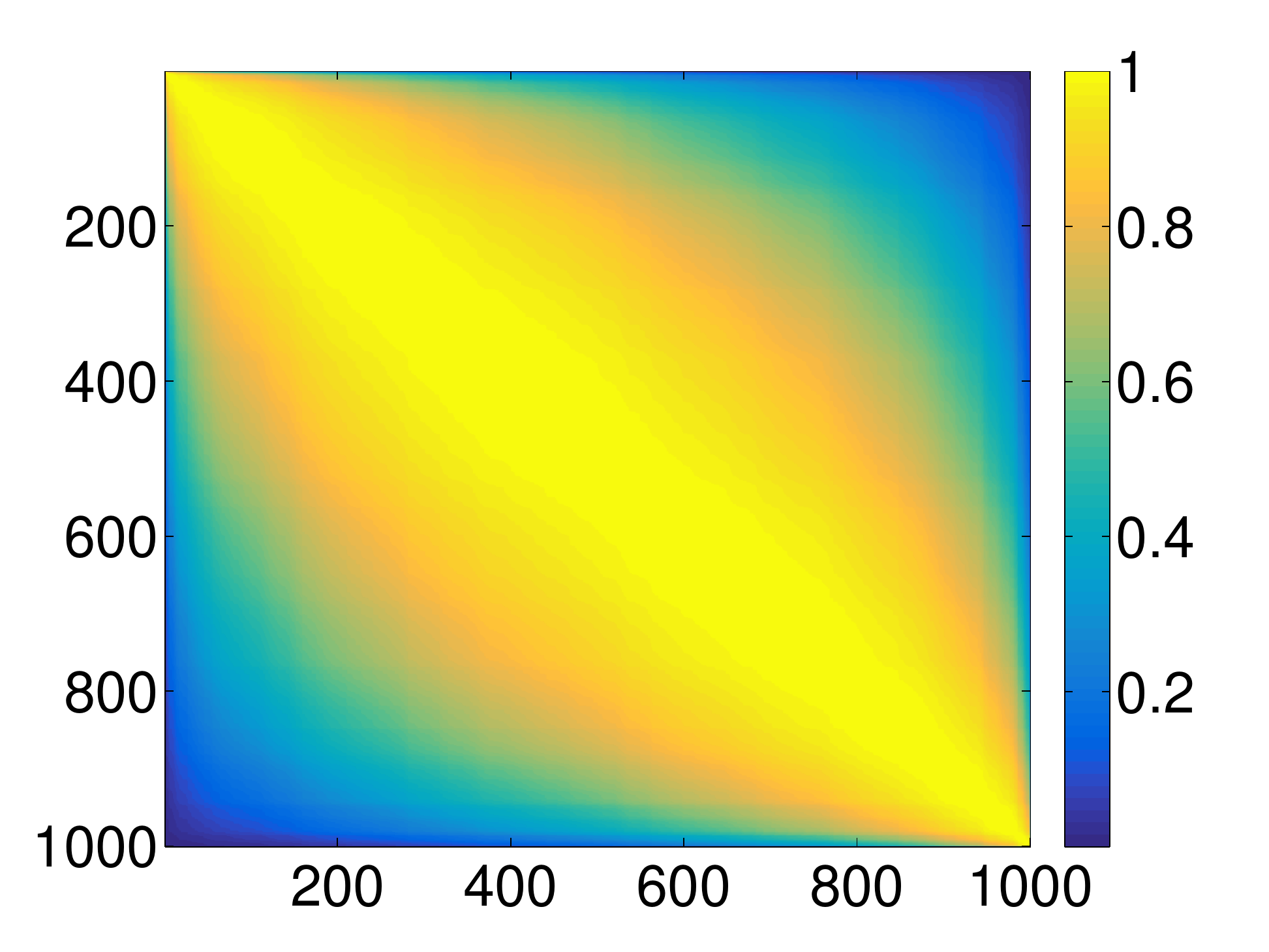}}
\subfigure[$K_\text{SKI}$ $(m = 40)$]{\label{fig:kski}\includegraphics[scale = 0.25]{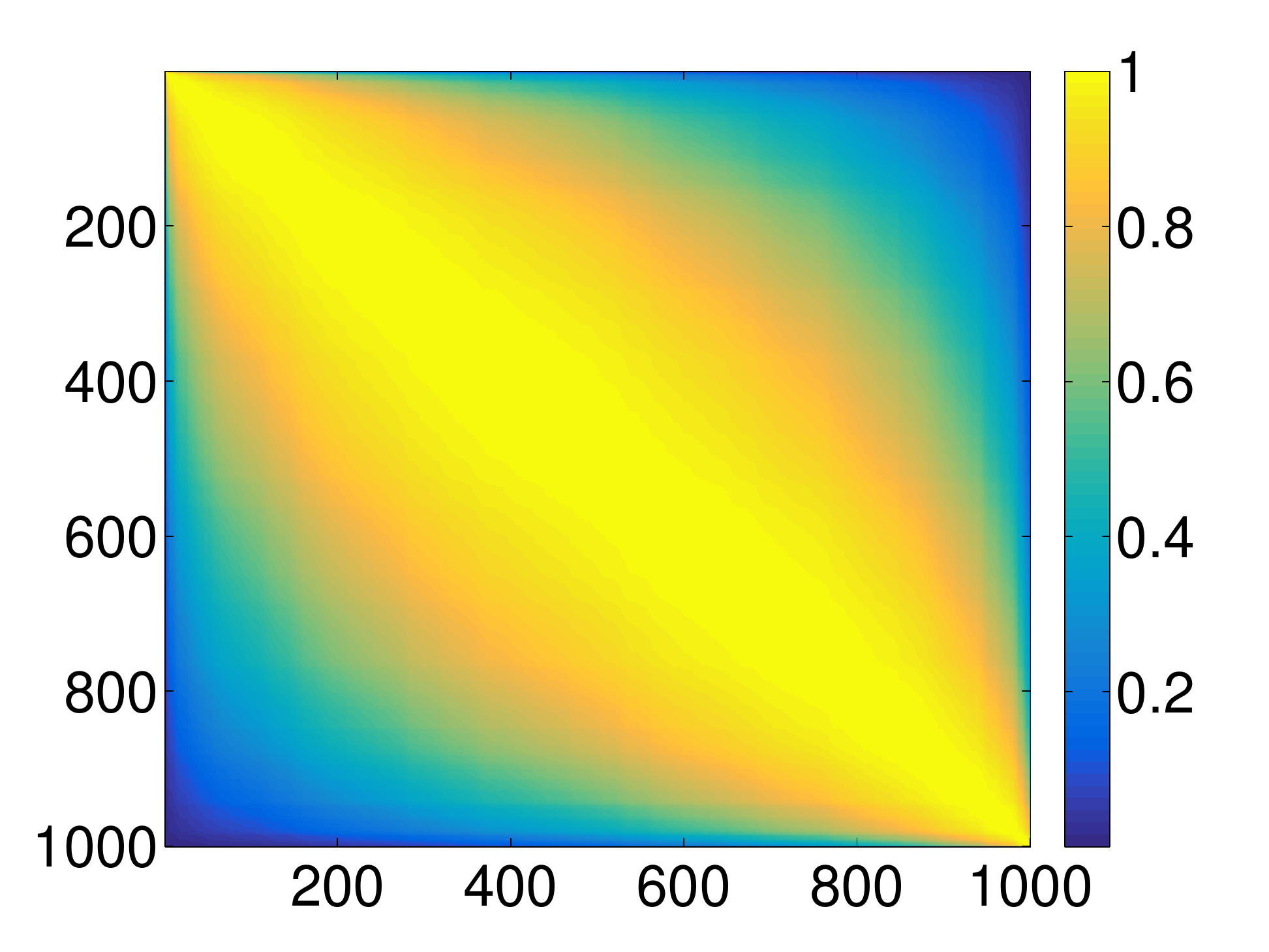}} 
\subfigure[$|K_\text{true} - K_\text{SKI, 40}|$]{\label{fig:err40}\includegraphics[scale = 0.25]{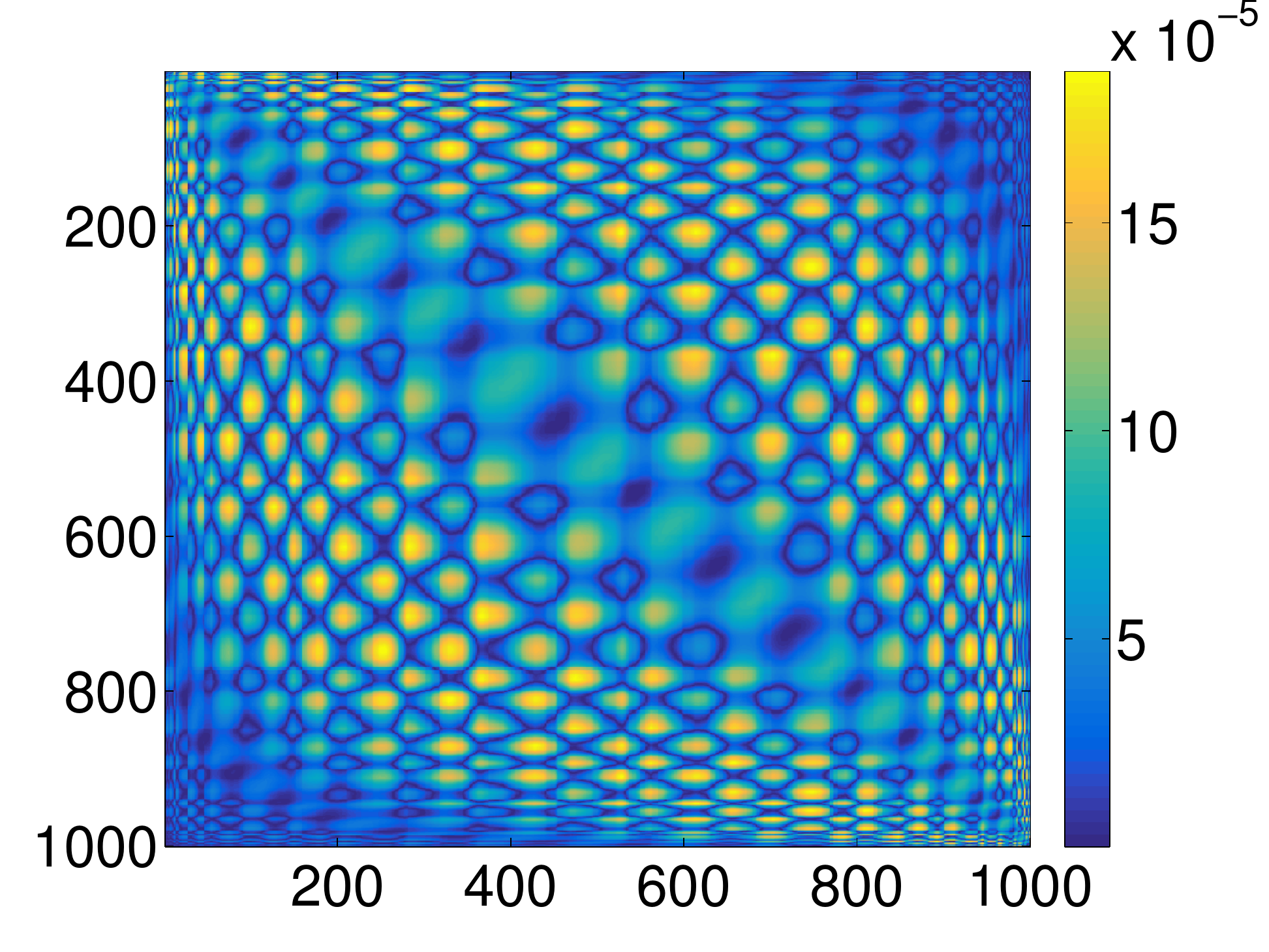}}
\subfigure[Interpolation Strategies]{\label{fig:varym}\includegraphics[scale = 0.25]{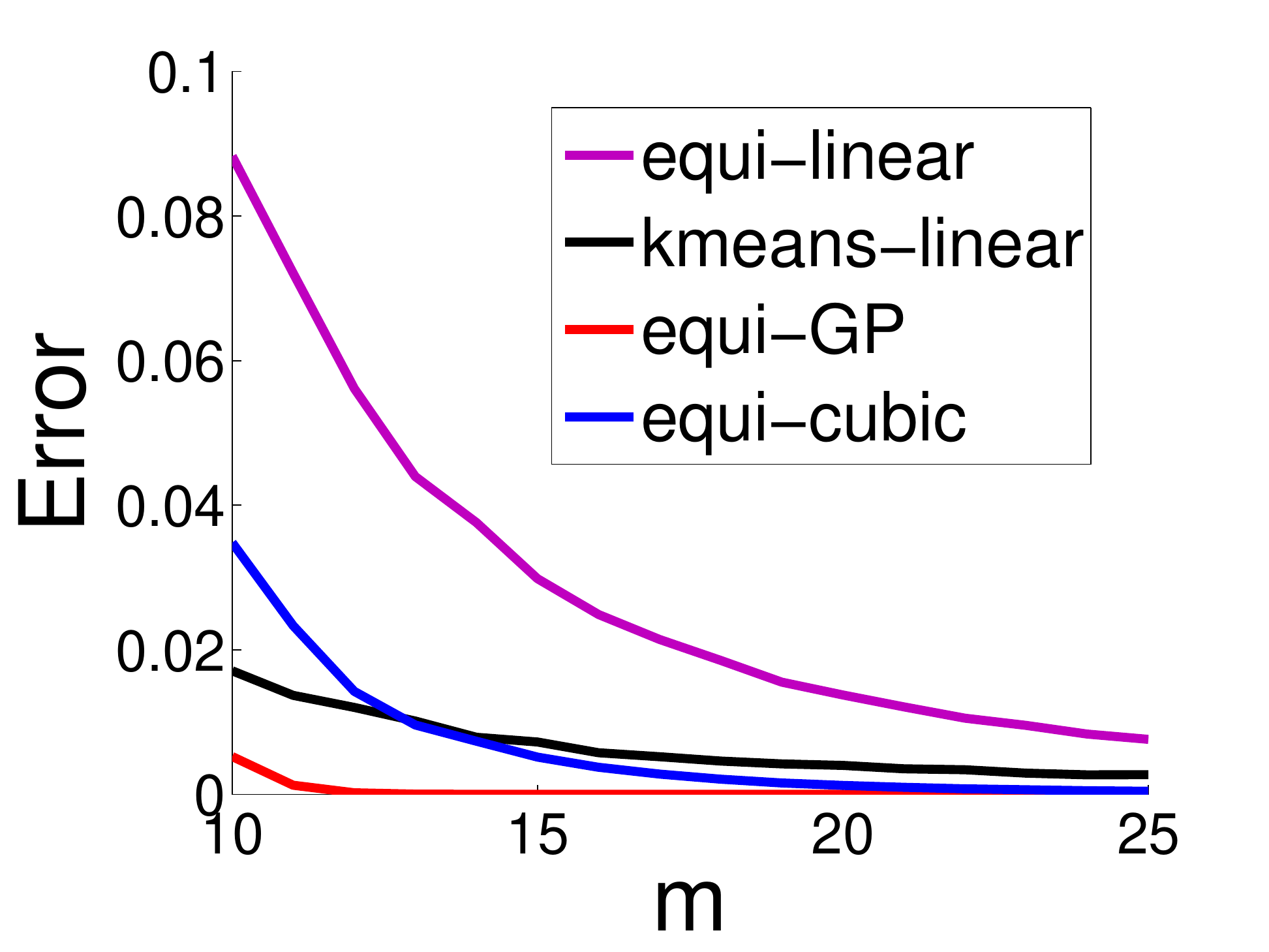}}  
\subfigure[$|K_\text{true} - K_\text{SKI, 150}|$]{\label{fig:ski150}\includegraphics[scale = 0.25]{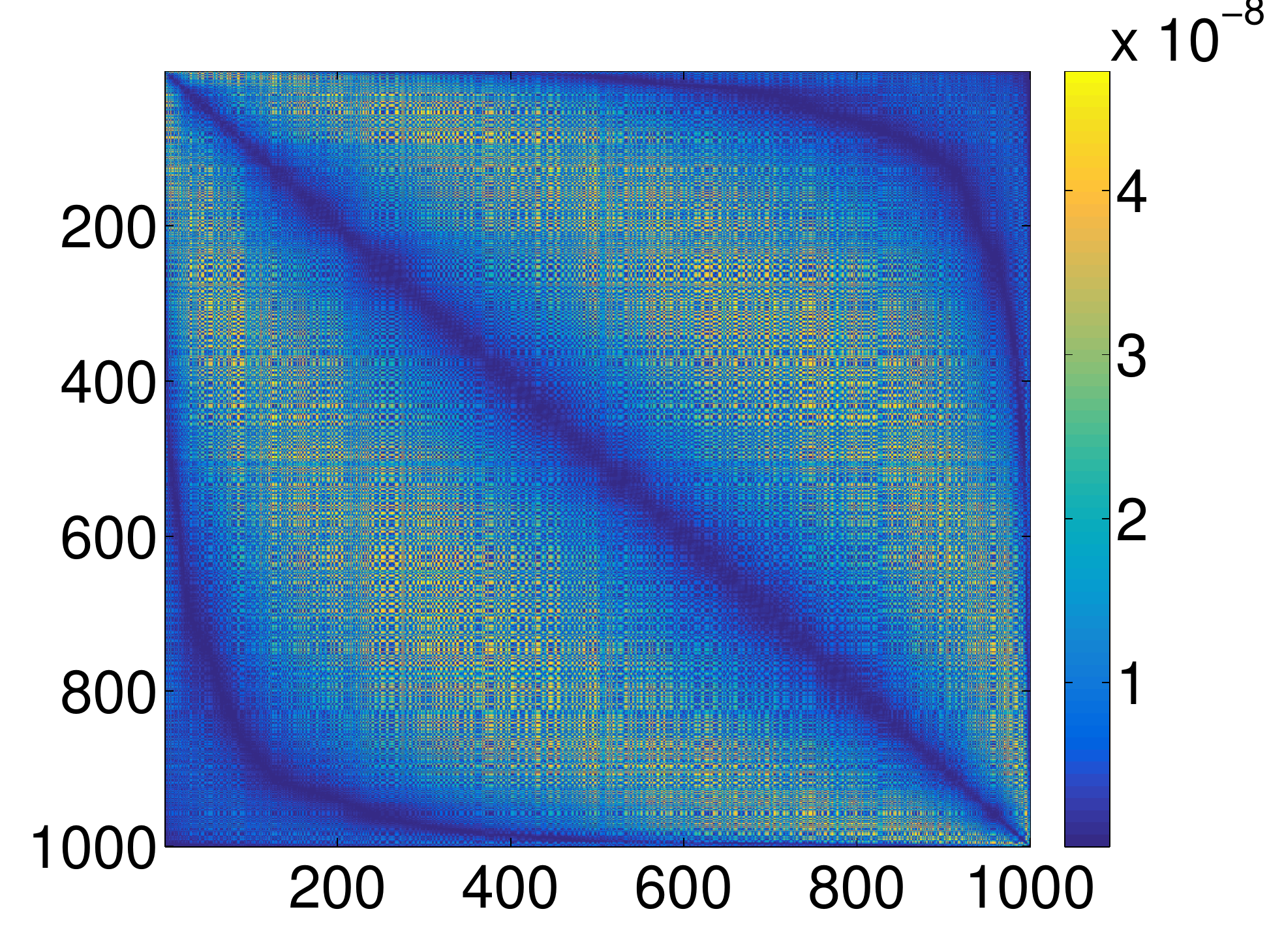}} 
\subfigure[$|K_\text{true} - K_\text{SoR, 150}|$]{\label{fig:sor150}\includegraphics[scale = 0.25]{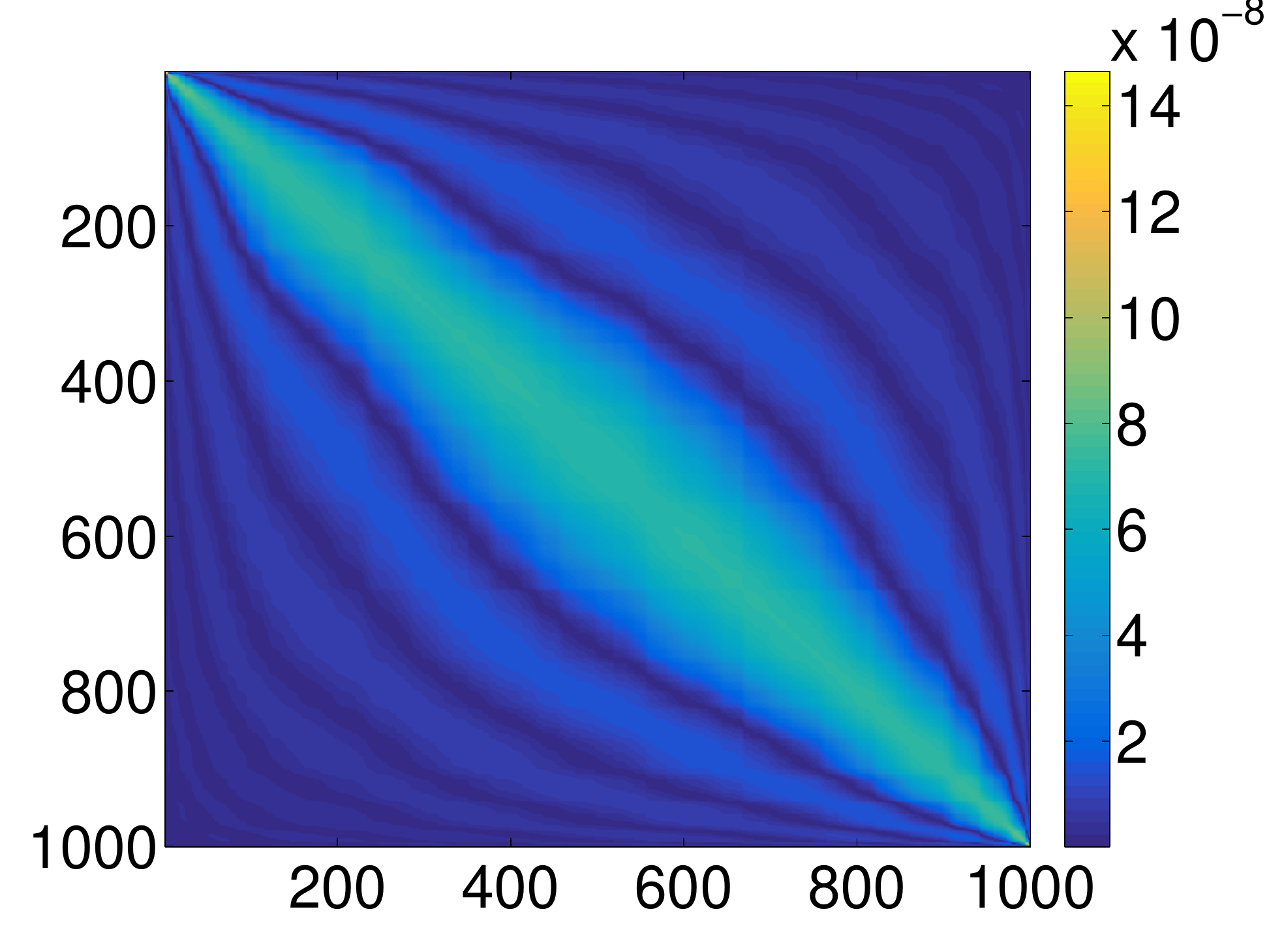}} 
\subfigure[Error vs Runtime]{\label{fig: runwin}\includegraphics[scale = 0.25]{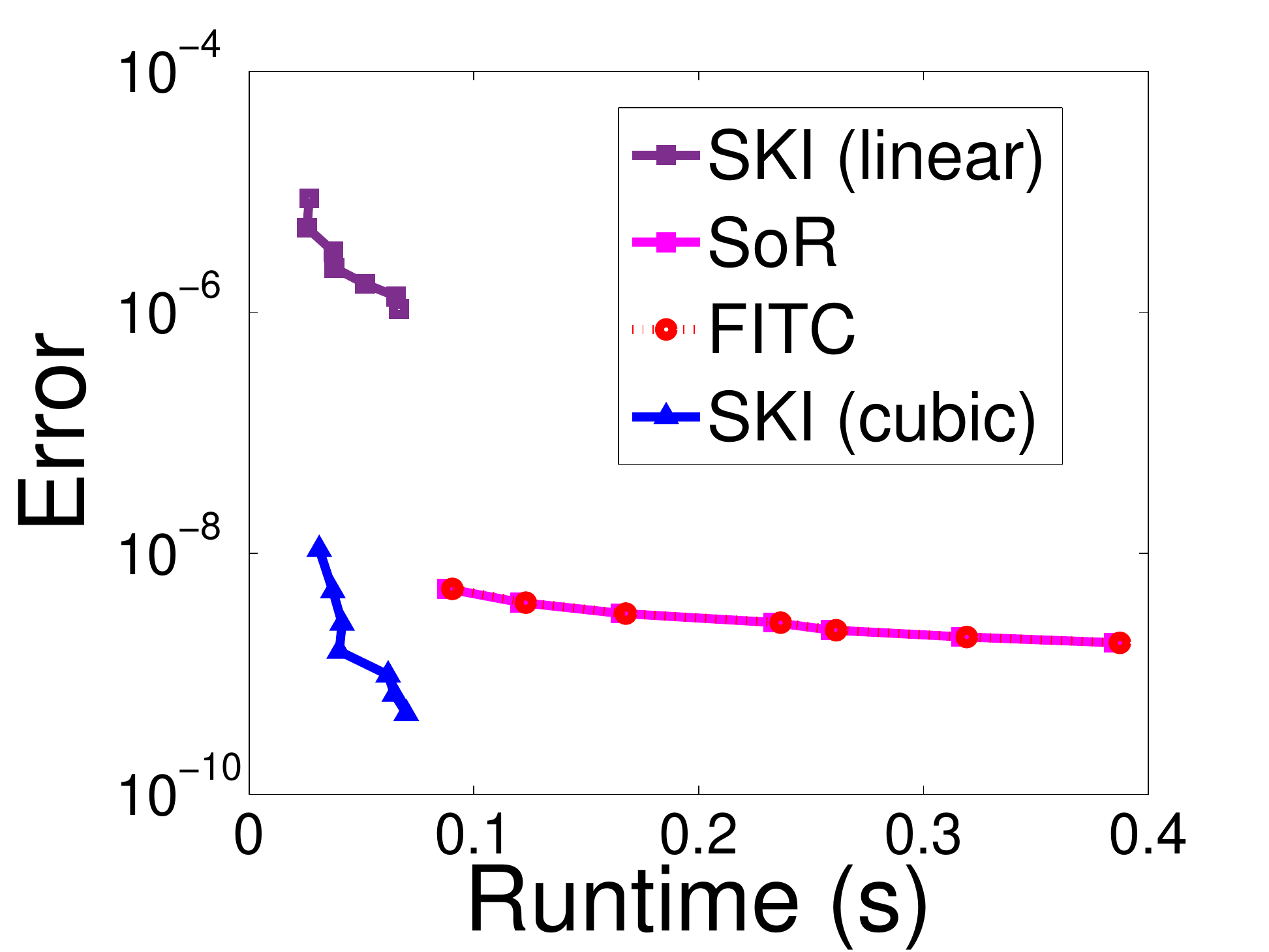}} 
\caption{Reconstructing a covariance matrix. a) True $1000 \times 1000$ RBF covariance matrix $K$.
b) $K_{\text{SKI}}$ reconstruction using local cubic interpolation and $m=40$ interpolation points. 
c) SKI absolute reconstruction error, for $m=40$.  
d) Average absolute error for reconstructing each entry of $K$ (the average of entries in (c)) as a function of $m$, using a regular grid with linear, cubic 
and GP interpolation (purple, blue, and red, respectively), and an irregular grid formed through $k$-means,
with inverse distance weighting interpolation (black).
e)-f) SKI (cubic) and SoR absolute reconstruction error, for $m=150$. g) Average absolute (log scale) error vs runtime, for $m \in [500, 2000]$.}
\label{fig: ktest}
\end{figure}

We generate a $1000 \times 1000$ covariance matrix $K$ from an RBF kernel 
(Eq.~\eqref{eqn: rbfcov}) evaluated 
at (sorted) inputs $X$ randomly sampled from $\mathcal{N}(0,25)$, shown in Figure~\ref{fig:ktrue}.  Note that the inputs have no 
grid structure.  The approximate $K$ produced by SKI using local cubic interpolation and only 40 interpolation points,
shown in Figure~\ref{fig:kski}, is almost indistinguishable from the original $K$.  Figure \ref{fig:err40} illustrates 
$|K - K_\text{SKI, m=40}|$, the absolute difference between the matrices in Figures \ref{fig:ktrue} and \ref{fig:kski}.
The approximation is generally accurate, with greatest precision near the diagonals and outer edges of $K$.

In Figure \ref{fig:varym}, we show how reconstruction error varies as a function of inducing points 
and interpolation strategy.  Local cubic and linear interpolation, using regular grids, are shown in 
blue and purple, respectively.  Cubic interpolation is significantly more accurate
for a small number of inducing points $m$.  In black, we also show the accuracy of using $k$-means
on the data inputs $X$ to choose inducing point locations.  In this case, we use local inverse distance
weighting interpolation, a type of linear interpolation which applies to irregular grids.  
This $k$-means strategy improves upon standard linear interpolation on a regular grid by 
choosing the inducing points which will be closest to the original inputs.  However, the value of using
$k$-means decreases when we are allowed more interpolation points, since the precise locations
of these interpolation points then becomes less critical, so long as we have general coverage of the 
input domain.  Indeed, except for small $m$, cubic interpolation on a regular grid generally outperforms 
inverse distance weighting with $k$-means.
Unsurprisingly, SKI with global GP kernel interpolation (shown in red), which corresponds
to the SoR approximation, is much more accurate than the other interpolation strategies for very small $m \ll n$.  

However, global GP kernel interpolation is much less efficient than local cubic kernel interpolation, and these accuracy
differences quickly shrink with increases in $m$.  Indeed in Figures  \ref{fig:ski150} and \ref{fig:sor150} we 
see both reconstruction errors are similarly small for $m=150$, but qualitatively different.  The error in
the SoR reconstruction is concentrated near the diagonal, whereas the error in SKI with cubic interpolation 
never reaches the top errors in SoR, and is more accurate than SoR near the diagonal, but is also more diffuse.
This finding suggests that combining cubic interpolation with GP interpolation could improve accuracy, if we
account for the regions where each is strongest.

Ultimately, however, the important question is not which approximation is most accurate for a given $m$, but 
which approximation is most accurate for a given runtime \citep{chalupka2013}.  In Figure \ref{fig: runwin} we compare the accuracies 
and runtimes for SoR, FITC, and SKI with local linear and local cubic interpolation, for $m \in [500, 2000]$ at $m=150$ unit increments.  
$m$ is sufficiently large that the differences in accuracy between SoR and FITC are negligible.  In general, the
difference in going from SKI with global GP interpolation (e.g.,\ SoR or FITC) to SKI with local cubic interpolation (KISS-GP)
is much more profound than the differences between SoR and FITC.  Moreover, moving from local linear interpolation to 
local cubic interpolation provides a great boost in accuracy without noticeably affecting runtime.  We also see that
SKI with local interpolation quickly picks up accuracy with increases in $m$, with local cubic interpolation actually surpassing 
SoR and FITC in accuracy for a given $m$.  Most importantly, for any given runtime, SKI with cubic interpolation is more accurate 
than the alternatives.

In this experiment we are testing the error and runtime for constructing
an approximate covariance matrix, but we are not yet performing inference with that covariance 
matrix, which is typically much more expensive, and where SKI will help the most.  Moreover, we are not 
yet using Kronecker or Toeplitz structure to accelerate SKI.  

\subsection{Kernel Learning}
\label{sec: klearn}

We now test the ability for SKI to learn kernels from data using Gaussian processes.  Indeed, SKI is intended 
to scale Gaussian processes to large datasets -- and large datasets provide a distinct opportunity to discover rich 
statistical representations through kernel learning.

Popular inducing point methods, such as FITC, improve the scalability of Gaussian processes.  However,
\citet{wilsonkernel2014} showed that these methods cannot typically be used for expressive kernel learning,
and are most suited to simple smoothing kernels.  In other words, these scalable GP methods often miss out 
on structure learning, one of the greatest motivations for considering large datasets in the first place.  This 
limitation arises because popular inducing methods require that the number of inducing points $m \ll n$, for 
computational tractability, which deprives us of the necessary information to learn intricate kernels.
SKI does not suffer from this problem, since we are free to choose large $m$; in fact, 
$m$ can be greater than $n$, while retaining significant efficiency gains over standard 
GPs.

\begin{figure}[!ht]
\centering
\subfigure{\label{fig: kone}\includegraphics[scale=0.3]{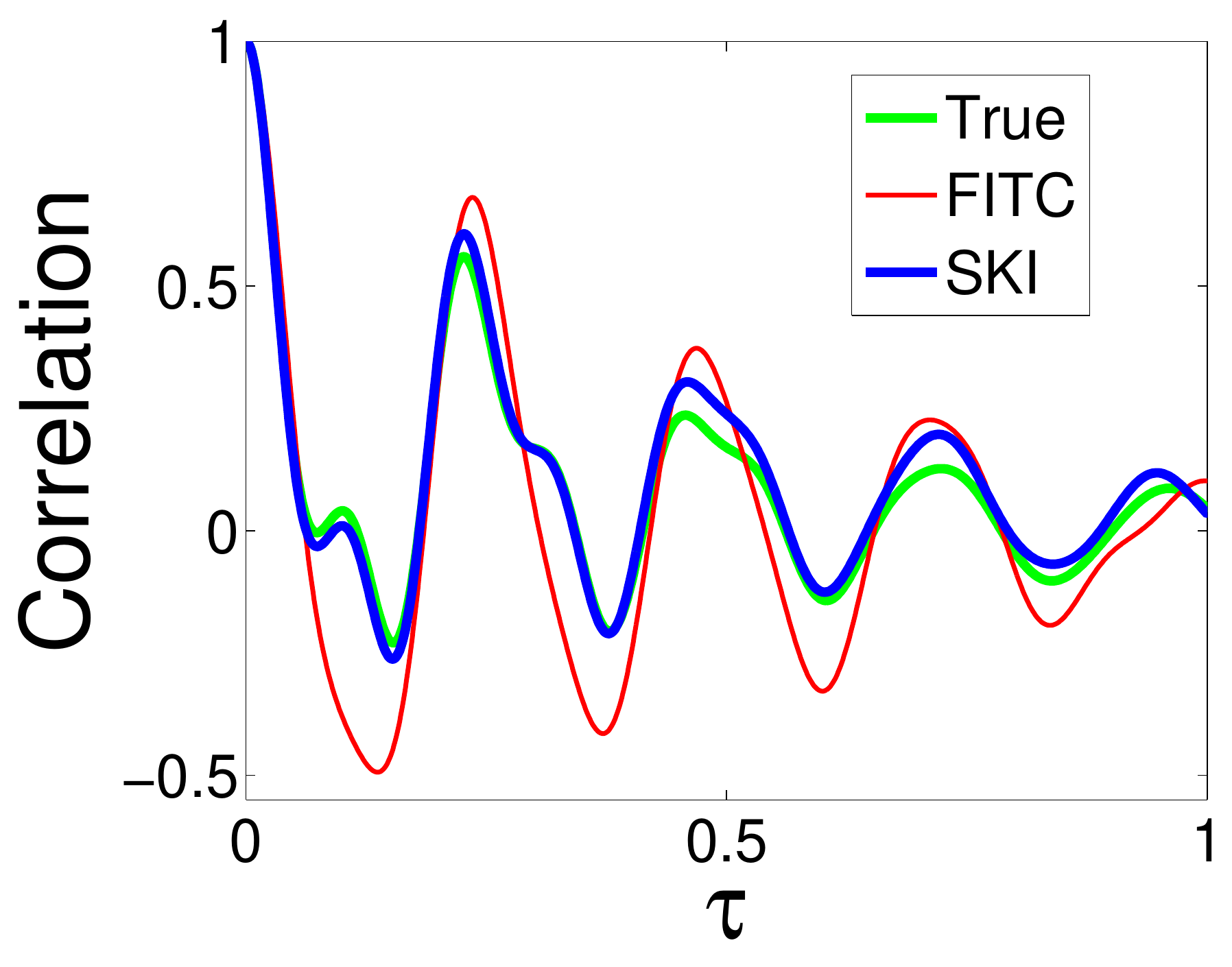}} 
\subfigure{\label{fig: ktwo}\includegraphics[scale=0.3]{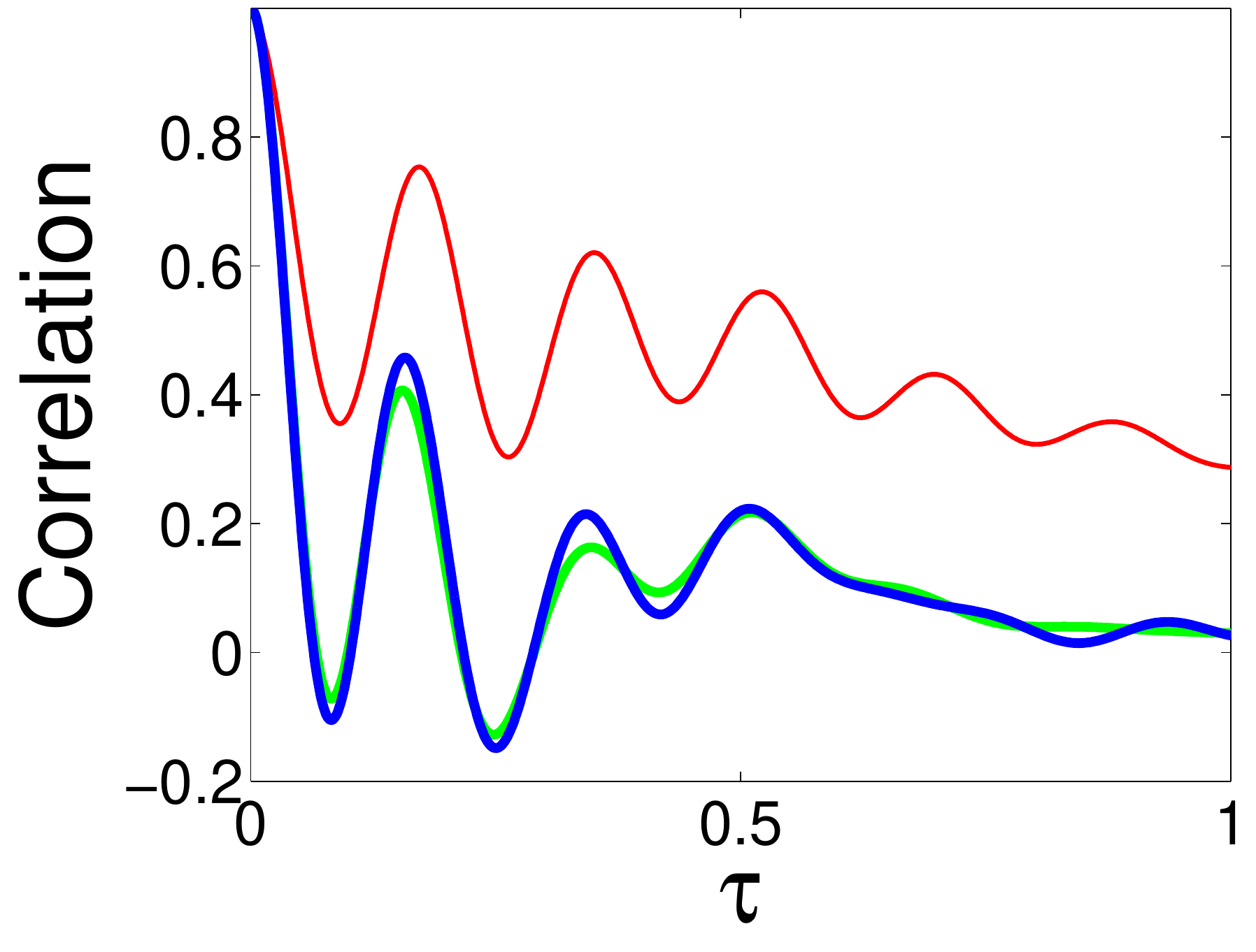}}
\caption{Kernel Learning.  A product of two kernels (shown in green) was used to sample
$10,000$ datapoints from a GP.  From this data, we performed kernel learning using SKI (cubic)
and FITC, with the results shown in blue and red, respectively.
All kernels are a function of $\tau = x-x'$ and are scaled by $k(0)$.}
\label{fig: klearnfig}
\end{figure}

To test SKI and FITC for kernel learning, we sample data from a GP which uses a known ground truth
kernel, and then attempt to learn this kernel from the data.  In particular, we sample $n = 10,000$ datapoints 
$\bm{y}$ from a Gaussian process with an intricate product kernel $k_{\text{true}}=k_1k_2$ queried at inputs $x \in \mathbb{R}^2$ drawn
from $\mathcal{N}(0,4I)$ (the inputs have no grid structure).  Each component kernel in the 
product operates on a separate input dimension, as shown in green in Figure ~\ref{fig: klearnfig}.
Incidentally, $n=10^4$ points is about the upper limit of what we can sample from a multivariate Gaussian distribution with a 
non-trivial covariance matrix.  Even a single sample from a GP with this many datapoints together with this 
sophisticated kernel
is computationally intensive, taking $1030$ seconds in this instance.  On the other hand, SKI can 
enable one to efficiently sample from extremely high dimensional ($n > 10^{10}$) non-trivial multivariate Gaussian 
distributions, which could be generally useful.\footnote{Sampling would proceed, e.g.,\ via $W_{\text{SKI}}[\text{chol}(K_1) \otimes \dots \otimes \text{chol}(K_p)] \nu$, \quad $\nu \sim \mathcal{N}(0,I)$.}

To learn the kernel underlying the data, we optimize the SKI and FITC marginal likelihoods of a Gaussian 
process $p(\bm{y} | \bm{\theta})$ with respect to the hyperparameters $\bm{\theta}$ of a spectral mixture 
kernel, using non-linear conjugate gradients.  In other words, the SKI and FITC kernels approximate a user
specified (e.g.,\ spectral mixture) kernel which is parametrized by $\bm{\theta}$, and to perform kernel learning 
we wish to learn $\bm{\theta}$ from the data.  Spectral mixture kernels \citep{wilsonadams2013} form a basis for all 
stationary covariance kernels, and are well-equipped for kernel learning.  For SKI, we use cubic interpolation and 
a $100 \times 100$ inducing point grid, equispaced in each input dimension.  That is, we have as many inducing 
points $m=10,000$ as we have training datapoints.   We use the same hyperparameter initialisation for each approach.

The results are shown in Figures \ref{fig: kone} and \ref{fig: ktwo}.  The true kernels are in green, the SKI reconstructions in blue, 
and the FITC reconstructions in red.  SKI provides a strong approximation, whereas FITC is unable 
to come near to reconstructing the true kernel.  In this multidimensional example, SKI leverages Kronecker 
structure for efficiency, and has a runtime of $2400$ seconds ($0.67$ hours), using $m = 10,000$ inducing 
points.  FITC, on the other hand, has a runtime of $2.6 \times 10^4$ seconds ($7.2$ hours), with only $m = 100$ inducing points.  
More inducing points with FITC breaks computational tractibility.  

Even though the locations of the training points are randomly sampled, in SKI we exploited the Kronecker structure in the 
covariance matrix $K_{U,U}$ over the inducing points $U$, to reduce the cost of using $10,000$ inducing points to less than
the cost of using $100$ inducing points with FITC.  FITC, and alternative inducing point methods, cannot effectively exploit 
Kronecker structure, because the non-sparse cross-covariance matrices $K_{X,U}$ and $K_{U,X}$ limit scaling to at best
$\mathcal{O}(m^2 n)$, as per section \ref{sec: SKI}.

\subsection{Natural Sound Modelling}
\label{sec: natsound}

In section \ref{sec: klearn} we exploited multidimensional Kronecker structure in the SKI covariance matrix $K_{U,U}$ for scalability.  
For targets indexed by a set of 
$1$D inputs, such as time series, we cannot exploit Kronecker structure for computational savings.  However, by placing the 
inducing points on a regular grid, we can create Toeplitz structure (section \ref{sec: toeplitz}) in $K_{U,U}$ which can be effectively 
exploited by SKI for additional scalability, but not by popular alternatives.

\begin{figure}[!ht]
\centering
\subfigure[Natural Sound]{\label{fig: sound}\includegraphics[scale=0.25]{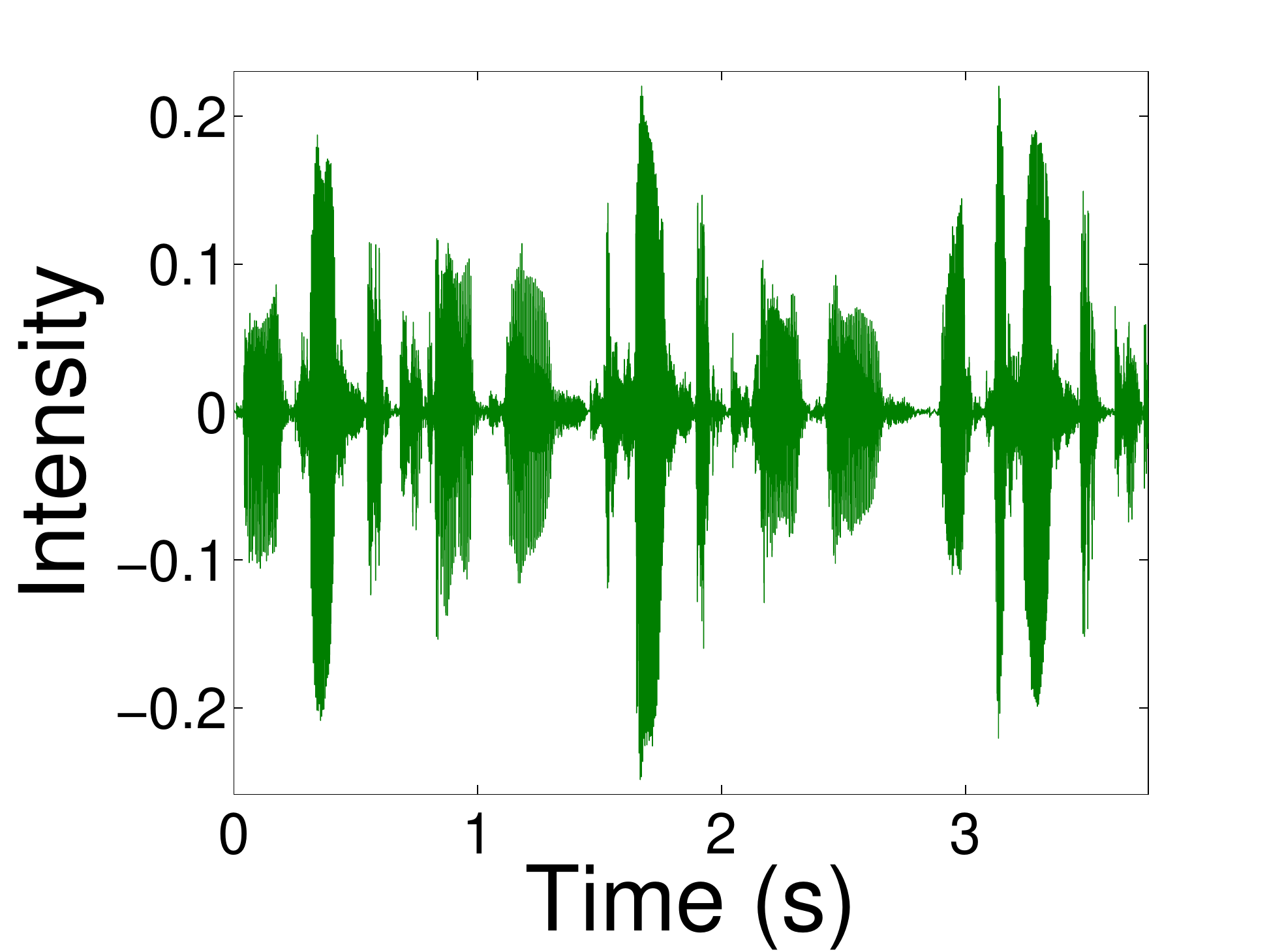}} 
\subfigure[Runtime vs $m$]{\label{fig: runtimem}\includegraphics[scale=0.25]{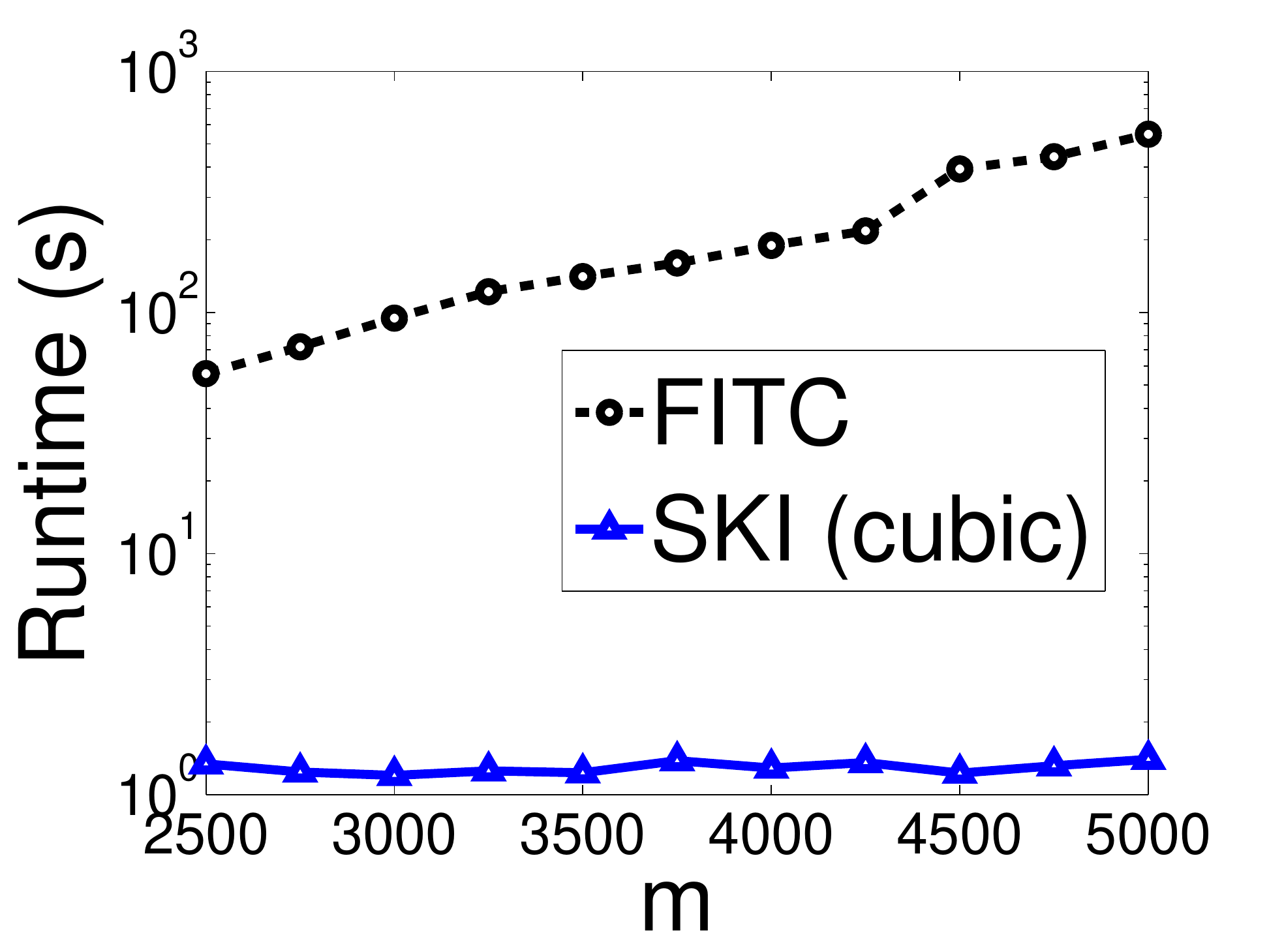}}
\subfigure[Error vs Runtime]{\label{fig: maetime}\includegraphics[scale=0.25]{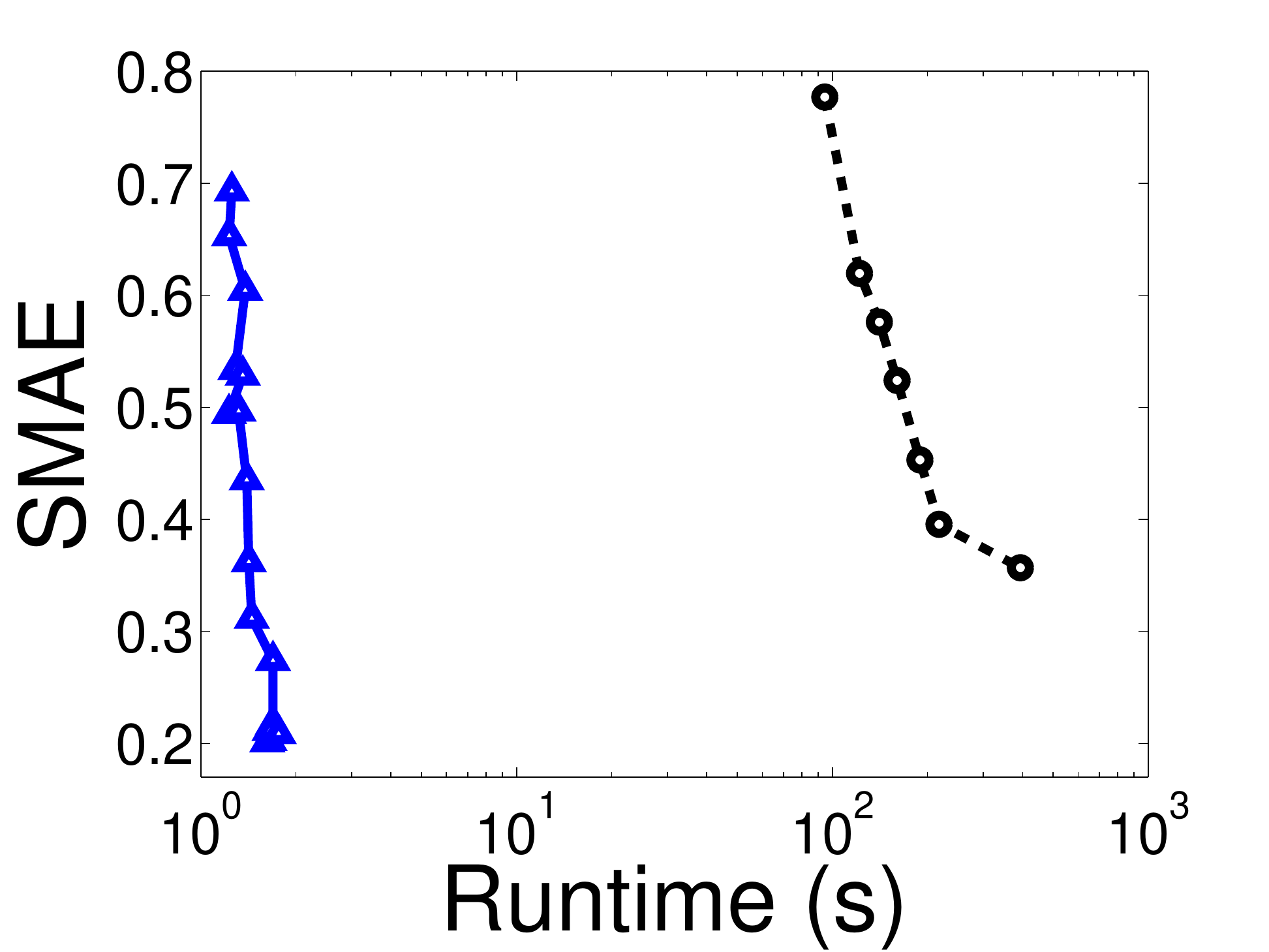}} 
\caption{Natural Sound Modelling.  We reconstruct contiguous missing regions of a natural
sound from $n = 59,306$ observations.  a) The observed data.  b) Runtime for SKI and FITC
(log scale) as a function of the number of inducing points $m$.  c) Testing SMAE error as a function of (log scale) 
runtime.}
\label{fig: natsoundfig}
\end{figure}

Gaussian processes have been successfully applied to natural sound modelling, with a view towards automatic speech recognition, 
and a deeper understanding of auditory processing in the brain \citep{turner10}.  
We use SKI to model the natural sound time series in Figure \ref{fig: sound}, considered in a different context by \citet{turner10}.  
We trained a full Gaussian process on a 
subset of the data, learning the hyperparameters of an RBF kernel, for use with both FITC and SKI.  We then used each of
these methods to reconstruct large contiguous missing regions in the signal. 
This time series does not have input grid structure
due to the high number of large arbitrarily located missing regions, and therefore direct
Toeplitz methods cannot be applied.  In total, there are 
$59,306$ training points and $691$ testing points. 
We place the inducing points for each method on a regular grid, and exploit Toeplitz structure in SKI for scalability.

Figure \ref{fig: runtimem} shows empirical runtimes (on a log scale) as a function of inducing points $m$ in both methods, and  
Figure \ref{fig: maetime} shows the standardised mean absolute error (SMAE) on test points as a function of runtime (log scale) for each 
method.\footnote{SMAE$_\text{method}$ = MAE$_\text{method}$ / MAE$_\text{empirical mean}$, so that the trivial solution of predicting 
with the empirical mean gives an SMAE of $1$, and lower values correspond to better fits of the data.}  For $m \in [2500,5000]$ the runtime
for SKI is essentially unaffected by increases in $m$, and hundreds of times faster than FITC, which does noticeably increase in runtime
with $m$.  Moreover, Figure \ref{fig: maetime} confirms our intuition that, for a given runtime, accuracy losses in going from GP kernel interpolation in FITC to
the more simple cubic kernel interpolation in the KISS-GP variant of SKI can be more than recovered by the gain in accuracy
enabled through more inducing points.  SKI has less than half of the error at less than 1\% the runtime cost of FITC.  SKI is generally able to infer 
the correct curvature in the function, while FITC, unable to use as many inducing points for any given runtime, tends to 
over-smooth the data.  Eventually, however, adding more inducing points increases runtime without increasing accuracy.  We also made predictions with SSGPR 
\citep{lazaro2010sparse}, a recent state of the art approach to scalable GP modelling, which requires $\mathcal{O}(m^2 n)$ computations 
and $\mathcal{O}(m^2)$ storage, for $m$ basis functions and $n$ training points.  For a range of $m \in [250, 1250]$, SSGPR had SMAE $\in$ $[1.12, 1.23]$ and 
runtimes $\in$ $[310, 8400]$ seconds.  Overall, SKI provides the best reconstruction of the signal at the lowest runtime.

\section{Discussion}
\label{sec: discussion}

We introduced a new \emph{structured kernel interpolation} (SKI) framework, which 
generalises and unifies inducing point methods for scalable Gaussian process 
inference.  In particular, we showed how standard inducing point methods correspond 
to kernel approximations formed through global Gaussian process kernel interpolation.
By changing to local cubic kernel interpolation, we introduced KISS-GP, a highly scalable
inducing point method, which naturally combines with Kronecker and Toeplitz 
algebra for additional gains in scalability.  Indeed we can view KISS-GP as relaxing 
the stringent grid assumptions in Kronecker and Toeplitz methods to arbitrarily located
inputs.  We showed that the ability for KISS-GP to efficiently handle a large number of
inducing points enabled expressive kernel learning and improved predictive accuracy, 
in addition to improved runtimes, over popular alternatives.  In particular, for any given
runtime, KISS-GP is orders of magnitude more accurate than the alternatives.  Overall, 
simplicity and generality are major strengths of the SKI framework.

We have only begun to explore what could be done with this new framework.  Structured
kernel interpolation opens the doors to a multitude of substantial new research directions.  For example,
one can create entirely new scalable Gaussian process models by changing interpolation
strategies.  These models could have remarkably different properties and applications.
And we can use the perspective given by structured kernel interpolation to better understand
the properties of any inducing point approach -- e.g.,\ which kernels are best approximated
by a given approach, and how many inducing points will be needed for good performance.
We can also combine new models generated from SKI with the orthogonal benefits of recent
stochastic variational inference for Gaussian processes. 
Moreover, the decomposition of the SKI covariance matrix into a Kronecker product of 
Toeplitz matrices provides motivation to unify scalable Kronecker and Toeplitz approaches.
We hope that SKI will inspire many new models and unifying perspectives, and an 
improved understanding of scalable Gaussian process methods.

\bibliographystyle{apalike}
\bibliography{mbibnew}

\end{document}